\pdfoutput=1

\documentclass[10pt,twocolumn,letterpaper]{article}

\usepackage[pagenumbers]{cvpr}

\usepackage{multirow}
\usepackage{array}
\usepackage{colortbl}
\newcolumntype{L}[1]{>{\raggedright\arraybackslash}p{#1}}
\usepackage{makecell}
\usepackage{enumitem}
\newcommand{\method}{\textup{\textsc{Skeptic}}\xspace}
\newcommand{\aspire}{ASPIRE\xspace}
\newcommand{\qwen}{Qwen3.8-27B\xspace}
\newcommand{\opus}{Claude Opus~5.5\xspace}

\newif\ifshowresults
\showresultstrue
\newcommand{\res}[1]{\ifshowresults\if\relax\detokenize{#1}\relax\textcolor{gray}{--}\else#1\fi\else\textcolor{gray}{--}\fi}
\newcommand{\yes}{\checkmark}

\usepackage{soul}
\definecolor{fighlred}{RGB}{255,200,200}
\definecolor{fighlyellow}{RGB}{255,236,160}
\definecolor{fighlgreen}{RGB}{196,236,196}
\DeclareRobustCommand{\fighl}[2]{{\sethlcolor{#1}\hl{#2}}}

\usepackage{placeins}

\definecolor{cvprblue}{rgb}{0.21,0.49,0.74}
\usepackage[pagebackref,breaklinks,colorlinks,allcolors=cvprblue]{hyperref}

\def\confName{CVPR}
\def\confYear{2027}

\title{Skeptic: Grounding Failure Diagnosis for Robot Skill Discovery}

\author{
Wenfang Sun$^{1}$\quad
Jianghao Deng$^{2}$\quad
Wenzhe Yin$^{1}$\quad
Yingjun Du$^{1\dagger}$\quad
Cees G.\,M. Snoek$^{1}$\\[0.6em]
$^{1}$University of Amsterdam\qquad
$^{2}$South China Normal University\\[0.6em]
{\normalsize\bfseries\href{https://wenfangsun.cn/Skeptic/}{Project Page \& Code}}
}

\begin{document}
\maketitle
{\let\thefootnote\relax\footnotetext{$^{\dagger}$Corresponding author.}}

\begin{abstract}
Robot agents discover reusable skills by revising programs from execution feedback and retaining experience for later tasks.
This process can fail when an agent treats an unsuccessful program as evidence that the task is physically impossible, ending exploration with revision budget remaining.
We study this failure in \qwen and examine two decisions that shape skill discovery: I) how execution failures are diagnosed, and II) which lessons enter the shared skill library.
Failure diagnosis requires the agent to reconstruct physical action effects from multimodal traces, while independent reproduction of a proposed repair does not require adequate task performance from its source program.
We introduce \method, a training-free protocol that uses measured action outcomes to guide repair and measured task success to govern skill admission.
A grounded event log reports action effects and object end states from depth images and proprioception, separating sensor observations from inferred diagnoses.
Verified promotion adds a development-success requirement on the source program to the existing independent reproduction check.
Across 60 audited tasks, the mean number of tasks with false impossibility judgments decreases from \res{19.5} to \res{8.5} per campaign.
On 80 LIBERO-Pro tasks and seven Robosuite tasks, \method improves held-out success by \res{14.9} and \res{11.7} percentage points, with frozen model weights and the same program-revision budget.
\end{abstract}

\section{Introduction}
\label{sec:intro}

Agentic skill discovery enables robots to improve control programs and accumulate reusable skills through interaction~\cite{liang2023cap,singh2023progprompt,huang2023voxposer,lu2026aspire}.
Its progress depends on how unsuccessful executions guide subsequent exploration.
An incorrect failure diagnosis can rule out a viable strategy and cause the agent to abandon a solvable task.
We use a frozen \qwen~\cite{qwen2026qwen38}, a 27B open-weight model, to study this problem in a robot-programming loop with persistent skill memory.
We focus on how evidence about an unsuccessful program becomes an unsupported judgment of physical impossibility, and how changing the evidence used for diagnosis and skill admission affects discovery.

\begin{figure}[t]
  \centering
  \includegraphics[width=\linewidth]{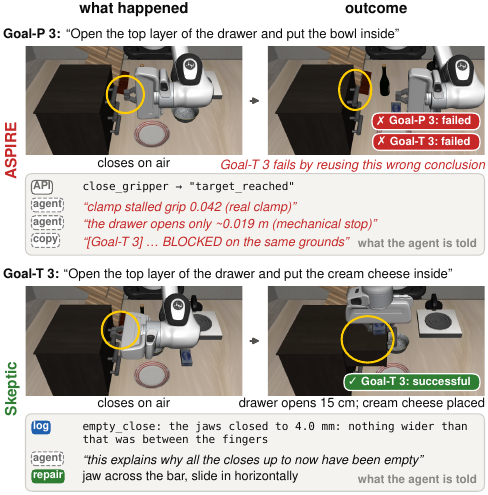}
  \caption{\textbf{A missed grasp becomes a claim of impossibility.}
  \emph{Top:} under \aspire~\cite{lu2026aspire}, \qwen closes the gripper on air; the API reports the closure as complete, and the agent concludes that the drawer is mechanically blocked, then reuses this conclusion on a second task with the same drawer.
  \emph{Bottom:} with \method, the event log reports the empty closure; the agent attributes the failure to its own grasp, changes how the gripper meets the handle, and completes that task.
  Quotes are from the run records; P and T denote position and task perturbation.}
  \label{fig:teaser}
\end{figure}

Physical feedback makes failure diagnosis difficult because completing an API call does not establish that the intended interaction occurred.
Coding agents receive images, primitive outputs, and execution traces~\cite{yuan2026unirobotagent,lu2026aspire}, from which they must reconstruct the physical effects of their programs before choosing a repair.
In \cref{fig:teaser}, the gripper closes without reaching the drawer handle, but the agent attributes the unsuccessful pull to a mechanical stop.
The coordinator then uses this explanation to close another task involving the same drawer.
The agent has generalized evidence about one unsuccessful interaction into a statement about task feasibility.
Such a conclusion can terminate exploration with revision budget remaining and constrain subsequent attempts.

We examine the distinction between using a strategy and discovering one on seven Goal tasks where \qwen initially fails every held-out episode.
Providing strategies and code sketches curated by \opus on these same tasks yields 342 successful episodes out of 350 (\cref{tab:supp-curation} in the supplement).
The model still adapts these recipes into task programs.
Rewriting or removing claims of impossibility from its own library and requiring three distinct strategies, however, yields only 9 successes out of 350.
The workers continue to explore local variations of their original approach while retaining the same explanation of failure.
On these tasks, the model can adapt an effective strategy but struggles to discover one through its own interaction.
Further attempts provide little progress when they are guided by the same mistaken explanation.

We introduce \method, a training-free protocol that uses measured action effects to guide program repair and measured task success to govern skill admission.
A \emph{grounded event log} organizes depth and proprioceptive measurements into action outcomes and object end states, distinguishing sensor observations from inferred diagnoses.
The worker can compare this record with its explanation of failure and the task instruction, locate an unfinished step, and preserve earlier actions supported by the execution record.
For skill accumulation, \emph{verified promotion} retains the existing independent reproduction check and adds a requirement on the source program of each proposed skill.
Reproducing a repair does not establish adequate task performance from the program that produced it; \method therefore requires the program to meet a measured development-success threshold before its proposal can enter the shared library.
An admission service computes success from execution records and checks the verifier's result before updating the library.

On all 80 LIBERO-Pro tasks~\cite{zhou2025liberopro}, \method raises the held-out success of \qwen from \res{63.4}\% to \res{78.3}\%, a gain of \res{14.9} percentage points, with the same model, starting library, and program-revision budget.
Applied unchanged to seven Robosuite tasks~\cite{zhu2020robosuite}, it increases mean success from \res{35.3}\% to \res{47.0}\%, a gain of \res{11.7} points.
Across the 60 audited Object, Goal, and Spatial tasks, the mean number of tasks with false impossibility judgments decreases from \res{19.5} to \res{8.5} per campaign.
Model weights remain frozen throughout skill discovery.

\section{Related Work}
\label{sec:related}

\noindent\textbf{Robot programming and agentic skill discovery.}
Vision-language-action models map visual observations and language instructions to robot actions~\cite{brohan2023rt2,octo2024,kim2024openvla,black2025pi0,pi2025pi05,nvidia2025gr00t}.
Programmatic approaches represent robot behavior as executable compositions of perception, planning, and control primitives~\cite{liang2023cap,singh2023progprompt,huang2023voxposer}, with affordance information guiding action selection~\cite{ahn2022saycan}.
This representation supports iterative improvement through execution and code revision.
CaP-X studies coding agents across models and interaction settings, demonstrating the benefits of multi-turn feedback and skill synthesis~\cite{fu2026capx}.
\aspire extends program repair with a shared library that transfers discovered strategies across tasks~\cite{lu2026aspire}.
Agentic world action models further connect reusable code with joint video and action prediction~\cite{liu2026aw0}.
World Action Agent rehearses proposed actions before execution~\cite{zhang2026waa}.
These developments motivate us in this work to study how compact open-weight models can acquire reusable skills through their own execution experience.

\noindent\textbf{Open-weight models for robot programming.}
Training-based approaches develop robot programming capabilities through supervised adaptation and reinforcement learning.
Octopus learns from generated training trajectories and environmental feedback~\cite{yang2024octopus}.
RoboCodeX trains multimodal code generation for robotic behavior synthesis~\cite{mu2024robocodex}, while Robo-Instruct uses simulator-checked programs and instruction alignment to fine-tune code models~\cite{hu2025roboinstruct}.
Guava distills frontier-model interaction trajectories into a compact agent~\cite{liu2026guava}; CaP-RL post-trains coding agents with verifiable rewards~\cite{fu2026capx}, and CaPtain combines supervised fine-tuning with multi-turn reinforcement learning~\cite{fan2026captain}.
Concurrent work post-trains 27B open-weight models in simulation to control the end effector directly: ZAP learns from the model's own successful rollouts and reinforcement learning within a minimal tool harness~\cite{pikus2026zap}, and Tang et al.\ train the model that we use with reinforcement learning alone, without a harness or a skill library~\cite{tang2026pear}.
Harness design also enables competitive robot manipulation with off-the-shelf open-weight models~\cite{yuan2026unirobotagent}.
SPARK composes a fixed library of reviewed skills into behavior trees, with the model that we use as planner~\cite{grant2026spark}.
We study skill discovery with fixed model weights, focusing on how execution feedback guides program repair and how experience accumulates across tasks.

\noindent\textbf{Execution feedback and failure diagnosis.}
Interpreting failed executions is central to autonomous robot improvement.
REFLECT constructs hierarchical summaries of multisensory robot experience and uses them to explain failures and generate correction plans~\cite{liu2023reflect}.
AHA trains a vision-language model to detect and reason about manipulation failures, providing feedback for downstream planning and control~\cite{duan2025aha}.
Coding-agent frameworks also expose execution traces, visual changes, and primitive-level feedback to support program revision~\cite{fu2026capx,lu2026aspire}.
Our study examines how a compact model uses this evidence during skill discovery, particularly when a failed interaction is misinterpreted as a physical constraint.
\method supplies measured action outcomes and object end states to help the agent locate the failed interaction and revise its program.

\noindent\textbf{Experience memory and verification.}
Persistent experience allows agents to reuse strategies discovered through interaction.
Lifelong robot library learning and self-evolving skill repositories accumulate reusable skills~\cite{tziafas2024lrll,xie2026uniskill}.
Verification already plays an important role: Voyager checks task completion before storing a skill~\cite{wang2023voyager}, and self-driven skill learning validates hypothesized subgoals through environment interaction before inducing skills~\cite{peng2024hvi}.
Reflexion, ExpeL, ACE, and ReasoningBank retain reflections, insights, or strategies derived from trajectories and outcome feedback~\cite{shinn2023reflexion,zhao2024expel,zhang2026ace,ouyang2026reasoningbank}.
Meta-Harness and RRSI further use evaluated candidates and execution experience to improve the agent harness itself~\cite{lee2026metaharness,xia2026rrsi}.
We examine a specific difficulty in robot skill discovery: an incorrect explanation of failure can terminate exploration and influence subsequent tasks.
\method connects failure diagnosis with skill admission, combining sensor-derived execution evidence with a measured source-attempt success requirement and independent reproduction before library updates.

\section{Method}
\label{sec:method}

\begin{figure*}[t]
  \centering
  \includegraphics[width=\textwidth]{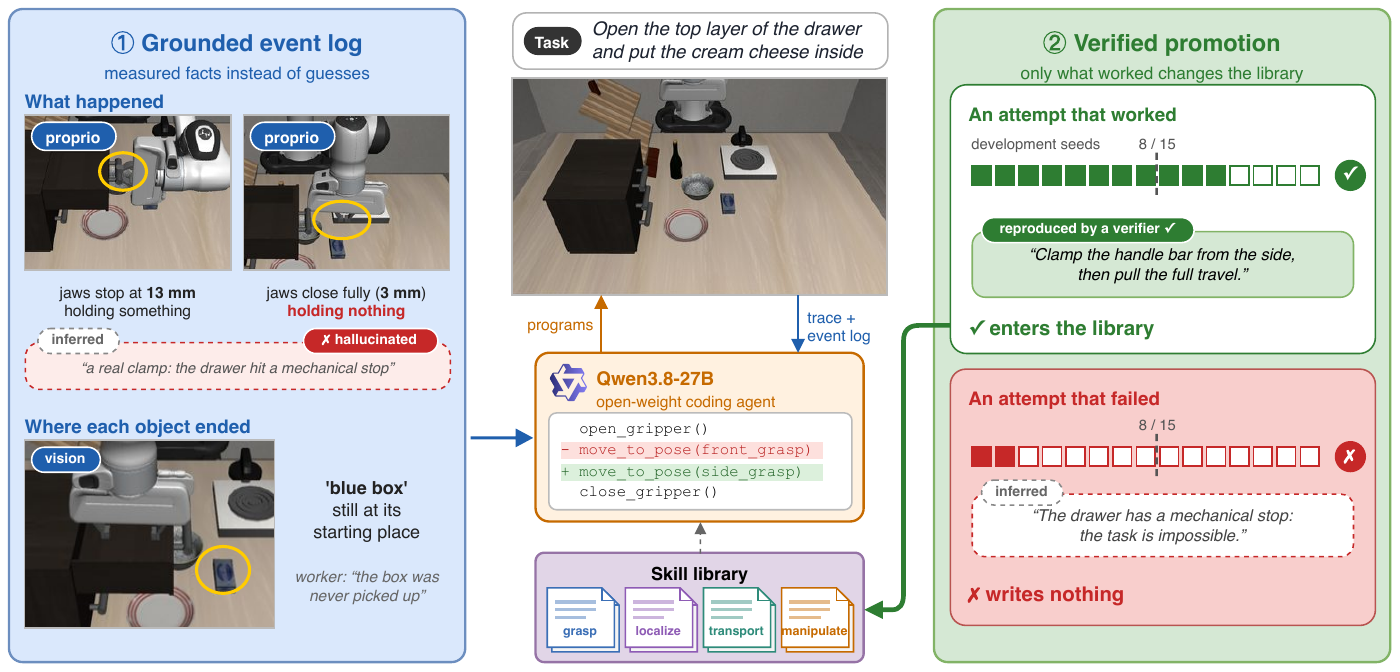}
  \caption{\textbf{Overview of \method.}
  Left: the grounded event log organizes execution measurements into action outcomes and object end states.
  Center: the coding agent uses this evidence to revise and execute robot programs.
  Right: a proposed skill enters the shared library when its source program meets a development-success threshold and an independent verifier reproduces the repair.
  On LIBERO-Pro, the threshold is at least 8 of 15 development seeds.}
  \label{fig:method}
\end{figure*}

\method connects program repair and skill accumulation to measured execution outcomes (\cref{fig:method}).
A grounded event log supplies evidence about the physical effects of an unsuccessful program.
Verified skill promotion determines whether a proposed repair is supported by both successful task execution and independent reproduction.
The two mechanisms operate within the existing discovery loop: the first informs the worker's next revision, while the second controls updates to the shared skill library.

\subsection{Robot programs and skill discovery}
\label{sec:background}

We follow the program-based discovery loop of \aspire~\cite{lu2026aspire}.
A campaign processes an ordered collection of manipulation tasks $\mathcal{T}$ using a shared skill library $\mathcal{L}$.
Each task $t$ specifies an instruction and a scene, with a simulator seed $\sigma$ determining the initial state.
A program $\pi$ follows the code-as-policy representation~\cite{liang2023cap}, composing fixed APIs for perception, grasp generation, motion planning, and control.
These APIs use SAM~3~\cite{carion2025sam3}, Contact-GraspNet~\cite{sundermeyer2021contactgraspnet}, and cuRobo~\cite{sundaralingam2023curobo}.
The agent develops its programs on a set of initial states indexed by development seeds $\mathcal{D}$.
For any seed set $\mathcal{S}$, program success is
\begin{equation}
  s_t(\pi,\mathcal{S})
  =
  \frac{1}{|\mathcal{S}|}
  \sum_{\sigma\in\mathcal{S}}
  \mathrm{succ}_t(\pi,\sigma),
  \label{eq:success}
\end{equation}
where $\mathrm{succ}_t(\pi,\sigma)$ is the simulator's binary task-completion predicate.

A coordinator dispatches a worker for each task.
The worker reads the instruction and current library, develops up to $K$ program versions using development executions, and returns a selected program $\pi_t^\ast$ together with proposed reusable repairs.
A fresh verifier receives each proposal's description and attempts to reproduce the repair.
Admitted proposals become entries in $\mathcal{L}$, providing in-context guidance for later tasks.
All three roles use the same \qwen model~\cite{qwen2026qwen38}, with weights fixed throughout the campaign.

\method supplies the worker with a grounded log after each failed development execution.
The worker uses this record alongside the original traces when revising its program.
When it proposes a skill, the proposal is associated with the program selected by that source attempt.
The promotion service then checks the program's development performance and the verifier's result before updating $\mathcal{L}$.
This association preserves the connection between a proposed lesson and the execution records that support its admission.

\subsection{Grounded execution feedback}
\label{sec:event-log}

The feedback must help the worker determine which interaction failed and what to change.
Primitive return values and images require interpretation: a completed motion or gripper call can leave the intended object unchanged.
\method organizes the available observations around the physical effects of these calls, making the evidence associated with each interaction explicit.

For a program $\pi$ executed from seed $\sigma$, the execution record contains primitive calls and their return values, robot proprioception at every simulation step, and agentview depth images captured at motion and gripper calls.
Objects in the log are those named by segmentation prompts already issued by the program.
From this record, \method constructs
$G_{\pi,\sigma}=(E_{\pi,\sigma},S_{\pi,\sigma})$,
where $E_{\pi,\sigma}$ is a chronological event sequence and
$S_{\pi,\sigma}$ summarizes the observed end states of the segmented objects.
Entries retain their source:
\texttt{[proprio]} denotes robot readings and API results,
\texttt{[vision]} denotes depth observations, and
\texttt{[inferred]} denotes diagnoses derived from those observations.
The worker can therefore inspect both a diagnosis and the measurements supporting it.

\smallskip
\noindent\textbf{Action outcomes.}
Each motion event records whether the hand reached its commanded target and the remaining distance when it stopped.
Each gripper event records the jaw opening and the visible surfaces within or near the region swept by the finger pads during closure.
These observations distinguish different outcomes that an API completion message alone leaves unresolved.
A motion shortfall directs attention to the approach, while jaw opening and nearby surfaces provide evidence about whether closure encountered an object.
Because entries are associated with the executed primitive calls, the worker can relate the observed discrepancy to the relevant operation in its program.

Depth changes in segmented regions provide additional evidence about object movement.
For example, the log may report that 20\% of the pixels in an object's initial region now show a farther surface.
The recorded fact is this depth change; interpretations such as lifting or slipping also use observations during subsequent hand motion and retain their \texttt{[inferred]} tag.
This preserves the distinction between the observed change and the physical explanation used to guide repair.

\Cref{fig:log} shows how these readings combine.
The jaws close to 4.0\,mm, while the depth observation shows no surface in the finger-pad closing region and a surface 2.6\,cm farther along the approach.
Together, these observations support an empty-closure diagnosis.
The worker can inspect the approach and grasp position using this evidence before attributing the unsuccessful interaction to a constraint on the object.

\begin{figure}[t]
  {\footnotesize\ttfamily\raggedright
  \#239-251 close\_gripper |
  \fighl{fighlyellow}{jaws closed to 4.0 mm}
  [proprio]\\[2pt]
  \quad |
  \fighl{fighlgreen}{no surface in the finger-pad closing region};
  nearest surface ahead:
  \fighl{fighlgreen}{+2.6 cm along the approach}
  [vision]\\[2pt]
  \quad !!
  \fighl{fighlred}{empty\_close: no object detected between the pads}
  [inferred]\\[2pt]
  end state | 'black bowl':
  \fighl{fighlgreen}{depth consistent with its starting location}
  [vision]\par}
  \caption{\textbf{A grounded event log} from a failed grasp of a bowl.
  Robot readings are highlighted in
  \fighl{fighlyellow}{yellow},
  depth observations in
  \fighl{fighlgreen}{green}, and the derived diagnosis in
  \fighl{fighlred}{red}.}
  \label{fig:log}
\end{figure}

\smallskip
\noindent\textbf{Object end states.}
Action events describe individual interactions, while repairing a multistep program also requires identifying which operation remains unfinished.
The log therefore ends with a summary for each segmented object.
It reports evidence that the object remains near its starting location or records where the hand last released it.
For a released object, the summary includes the release point relative to other segmented objects' initial footprints and the surface height observed there after the hand moves away.
These observations connect the action sequence to the spatial arrangement left by the program.

The worker compares this summary with the task instruction.
Consider a task requiring two objects to be placed in a basket.
If one object was released over the basket's initial footprint while the other remains at its starting location, the record directs attention to the second pick-and-place sequence.
The worker can inspect that sequence before changing the earlier operations.
The summary thus provides evidence for choosing the scope of a revision: which interaction needs further investigation and which earlier actions remain supported by the execution record.
The log reports observed events and spatial relationships; the worker determines their relevance to the requested task.

\smallskip
\noindent\textbf{Feedback delivery.}
A fixed wrapper writes the log beside the original results of each failed development run.
When a batch of executions finishes, the runner prints a one-line summary for each failed seed, including its object end states.
The worker can compare outcomes across initial states, identify the operations associated with failure, and return to the detailed event sequence when modifying the program.
The logger constructs this feedback from the records of those executions, without requiring another robot rollout or consuming a program revision.

\subsection{Verified skill promotion}
\label{sec:promotion}

The event log informs repair during an attempt.
Skill promotion determines whether the resulting experience becomes guidance for subsequent tasks.
Under \aspire, an independent verifier attempts to reproduce a proposed repair from its description~\cite{lu2026aspire}.
This checks whether the repair can be reproduced, while leaving the task performance of its source program outside the admission condition.
\method adds that source-program performance as a second requirement.

For a skill $c$ proposed during task $t$, let $\pi_c$ denote the program selected by the attempt that produced the proposal, and let $V(c)$ denote the verifier's verdict.
The original and revised admission rules are
\begin{align}
  \mathrm{admit}_{\aspire}(c)
  &=
  [\,V(c)=\text{sound}\,],
  \label{eq:aspire-admit}\\
  \mathrm{admit}(c)
  &=
  [\,s_t(\pi_c,\mathcal{D})\geq\tau\,]
  \wedge
  [\,V(c)=\text{sound}\,].
  \label{eq:admit}
\end{align}
The added condition requires evidence that the source program completes the task across development initial states.
On LIBERO-Pro, $|\mathcal{D}|=15$ and $\tau=8/15$, so the program must succeed on at least eight development seeds.

The two checks evaluate different aspects of a proposal.
Independent reproduction tests whether another agent can implement the repair from its description.
Development success ties that repair to an originating program with measured task performance.
The admission decision combines both pieces of evidence before making the proposal available for reuse.

The success requirement applies to the selected program across $\mathcal{D}$.
Individual failed executions continue to generate logs and inform program revision.
As development proceeds, a revised program can reach the threshold and support a skill proposal.
A promoted entry may describe both its working strategy and unsuccessful configurations encountered during development.
Its eligibility rests on the associated program's measured performance and the verifier's reproduction.

Admission is enforced by a separate promotion service.
The service computes the selected program's success rate from its actual run records, reads the independent verifier's verdict, and updates $\mathcal{L}$ only when both conditions hold.
Agents access the library as read-only.
The logger and promotion service operate outside the worker's editable programs, so the records used for repair and admission are supplied by the execution infrastructure.
Newly admitted skills then become guidance for later workers within the same discovery campaign.

\section{Experiments}
\label{sec:experiments}

We test whether measured execution feedback and verified skill promotion help a frozen open-weight agent discover successful robot programs.

\subsection{Experimental setup}
\label{sec:setup}

\noindent\textbf{Controlled comparison.}
We compare \method with the original \aspire protocol using the same \qwen model, robot APIs, released starting library, seeds, and program-revision budget.
For LIBERO-Pro, the model runs in FP8 through vLLM~\cite{kwon2023vllm} on a node with four H100 GPUs, which also hosts perception, motion planning, and eight simulation slots.
Each campaign covers the 60 Object, Goal, and Spatial tasks followed by the 20 Long tasks.
Following \aspire's backward skill transfer~\cite{lu2026aspire}, once all tasks are finished, tasks below 80\% held-out success are rerun with the complete skill library.
Both protocols apply this rule unchanged, and we report success on 50 held-out seeds per task after the rerun.

\smallskip
\noindent\textbf{Reporting.}
The main comparison averages two complete campaigns per protocol; a campaign runs the task sequence and accumulates its own skill library.
We average each task's scores across campaigns before averaging over tasks.
Paired bootstrapping over tasks with 10{,}000 resamples gives 95\% confidence intervals for success differences.
These intervals capture task sampling; we also report campaign scores to show variation across repeated discovery runs.

\begin{table*}[t]
  \centering
  \showresultstrue
  \setlength{\tabcolsep}{4pt}
  \begin{tabular*}{\textwidth}{@{\extracolsep{\fill}}l l cc cc cc cc c c@{}}
    \toprule
    & & \multicolumn{2}{c}{Object} & \multicolumn{2}{c}{Goal} & \multicolumn{2}{c}{Spatial} & \multicolumn{2}{c}{Long} & & \\
    \cmidrule(lr){3-4}\cmidrule(lr){5-6}\cmidrule(lr){7-8}\cmidrule(lr){9-10}
    Model & Protocol & Pos & Task & Pos & Task & Pos & Task & Pos & Task & All & API \$/task \\
    \midrule
    \multicolumn{12}{@{}l}{\textit{Proprietary models, reported by \aspire~\cite{lu2026aspire}}} \\
    Claude Opus 4.6 & \aspire & 98.0 & 95.0 & 47.0 & 42.0 & 42.0 & 47.0 & 43.2 & 52.0 & 58.3 & 22.3 \\
    Claude Opus 5 & \aspire & 98.6 & 100.0 & 97.0 & 86.8 & 97.8 & 98.0 & 81.0 & 83.4 & 92.8 & 39.1 \\
    GPT-6 Astra & \aspire & 99.0 & 99.4 & 97.0 & 96.8 & 97.0 & 97.6 & 80.8 & 95.0 & 95.3 & 18.1 \\
    \opus & \aspire & 98.2 & 100.0 & 97.0 & 99.4 & 99.4 & 99.2 & 97.0 & 98.8 & 98.6 & 12.4 \\
    \midrule
    \multicolumn{12}{@{}l}{\textit{Open-weight model, our runs}} \\
    \qwen & \aspire & \res{88.0} & \res{79.2} & \res{50.9} & \res{52.8} & \res{67.2} & \res{82.4} & \res{36.7} & \res{50.0} & \res{63.4} & \res{0} \\
    \rowcolor{blue!4}\qwen & \method & \res{99.6} & \res{98.5} & \res{70.9} & \res{63.8} & \res{91.5} & \res{96.8} & \res{43.8} & \res{61.4} & \res{78.3} & \res{0} \\
    \bottomrule
  \end{tabular*}
  \caption{\textbf{Held-out success (\%) on all 80 LIBERO-Pro tasks.}
  Each Pos/Task column contains 10 tasks evaluated on 50 seeds each; Pos and Task denote position and task perturbations.
  All is the mean of the eight columns.
  Proprietary-model results are from \aspire's per-suite tables, using iterative debugging without evolutionary search~\cite{lu2026aspire}; API charges combine the 60-task and Long runs at list prices.
  Our two rows each average two complete campaigns.
  Local inference incurs no API charges; development resources are reported in \cref{fig:compute}.}
  \label{tab:main}
\end{table*}

\subsection{Main results}
\label{sec:results}

\noindent\textbf{LIBERO-Pro success.}
\method raises held-out success from \res{63.4}\% to \res{78.3}\% across all 80 tasks, a \res{14.9}-point gain (95\% CI \res{[8.4, 21.6]}; \cref{tab:main}).
On the 60 Object, Goal, and Spatial tasks, success rises from \res{70.1}\% to \res{86.9}\% (\res{+16.8} points, 95\% CI \res{[9.1, 24.8]}).
Both \method campaigns improve over both baseline campaigns on these 60 tasks: their scores are \res{89.2}\% and \res{84.5}\%, compared with \res{63.6}\% and \res{76.6}\%.
The lower-scoring \method campaign exceeds the higher-scoring baseline by \res{7.9} points.

\smallskip
\noindent\textbf{Differences across task suites.}
Object and Spatial tasks mainly require grasping and placement; their suite means improve by \res{15.5} and \res{19.4} points, respectively.
All four Object/Spatial columns finish within \res{8} points of \opus.
Goal tasks also improve by \res{15.5} points, but drawers, knobs, and pushing leave a larger gap to the frontier model.
Long tasks improve by \res{9.3} points.
The pattern fits the feedback supplied by the log: missed grasps and incomplete reaches can be measured directly, whereas successful contact manipulation may still require a strategy the agent has not proposed.

\smallskip
\noindent\textbf{Remaining failures.}
Using the two-campaign means, eleven of the 60 Object, Goal, and Spatial tasks remain below 80\%, ten of them Goal tasks.
Pushing the plate to the front of the stove (Goal-P 5) and opening the bottom drawer (Goal-T 0) fail in every reported 27B campaign, with no successful development seed.
The other nine reach at least \res{86}\% in one of the four campaigns in \cref{tab:main}.
Thus, many residual failures still depend on the discovery run: the same model finds a successful program in one campaign and misses it in another.

\begin{table}[t]
  \centering
  \showresultstrue
  \small
  \setlength{\tabcolsep}{4pt}
  \begin{tabular}{@{}l cc c>{\columncolor{blue!4}}c@{}}
    \toprule
    & CaP-X & \aspire & \multicolumn{2}{c}{\qwen} \\
    \cmidrule(lr){2-2}\cmidrule(lr){3-3}\cmidrule(l){4-5}
    Task & gpt-oss-120B & Opus 5.5 & \aspire & \method \\
    \midrule
    Cube lift & 73 & 100 & \res{100} & \res{100} \\
    Cube stack & 51 & 100 & \res{85} & \res{91} \\
    Cube restack & 38 & 100 & \res{62} & \res{97} \\
    Spill wipe & 96 & 100 & \res{0} & \res{4} \\
    2-arm handover & 0 & 100 & \res{0} & \res{8} \\
    2-arm lift & 16 & 100 & \res{0} & \res{29} \\
    Nut assembly & 0 & 100 & \res{0} & \res{0} \\
    \midrule
    Mean & 39 & 100 & \res{35.3} & \res{47.0} \\
    \bottomrule
  \end{tabular}
  \caption{\textbf{Held-out success (\%) on seven Robosuite tasks}, with 100 seeds per task.
  gpt-oss-120B results are from CaP-X~\cite{fu2026capx} (tier M1); Claude Opus~5.5 results are from \aspire~\cite{lu2026aspire}.
}
  \label{tab:other}
\end{table}

\smallskip
\noindent\textbf{Transfer to Robosuite.}
\label{sec:other}
We apply the same rules to seven tasks~\cite{zhu2020robosuite}: five single-arm and two bimanual.
Both protocols develop programs on seeds 101--125 and evaluate on 1--100; promotion requires at least 14 of the 25 development seeds.
We remove the released finished programs for these tasks; the remaining starting library is as on LIBERO-Pro.
The log uses proprioception and depth images, with one log per arm on bimanual tasks.
Mean success increases from \res{35.3}\% to \res{47.0}\% (\cref{tab:other}).
\method is higher on five of seven tasks and lower on none.
Most of the gain comes from cube restacking (\res{62}\% to \res{97}\%) and two-arm lifting (\res{0}\% to \res{29}\%).
Cube stacking improves from \res{85}\% to \res{91}\%, two-arm handover from \res{0}\% to \res{8}\%, and spill wiping from \res{0}\% to \res{4}\%; cube lifting stays at \res{100}\%.

\smallskip
\noindent\textbf{Failures that persist across simulators.}
Nut assembly stays at \res{0}\% under both protocols, and spill wiping and two-arm handover stay below \res{10}\% under \method.
Nut assembly reproduces the failure studied in this paper: under \method, the worker still declares the task physically infeasible after its failed attempts.
Verified promotion excludes the failed attempt's conclusion from the library, but the log still leaves room for a worker to misinterpret failure and stop searching.

\subsection{Ablations and failure analysis}
\label{sec:ablation}

\begin{table}[t]
  \centering
  \showresultstrue
  \small
  \setlength{\tabcolsep}{2.0pt}
  \begin{tabular}{@{}l ccc ccc cc@{}}
    \toprule
    & {\scriptsize\makecell{Verified\\promotion}} & {\scriptsize\makecell{Action\\events}} & {\scriptsize\makecell{End\\states}} & Object & Goal & Spatial & Final & First \\
    \midrule
    \aspire & & & & \res{83.6} & \res{51.9} & \res{74.8} & \res{70.1} & \res{61.3} \\
    & \yes & & & \res{92.0} & \res{68.2} & \res{83.6} & \res{81.3} & \res{54.0} \\
    & & \yes & \yes & \res{91.1} & \res{60.2} & \res{91.1} & \res{80.8} & \res{72.2} \\
    & \yes & \yes & & \res{94.7} & \res{63.6} & \res{92.6} & \res{83.6} & \res{76.9} \\
    \rowcolor{blue!4}\method & \yes & \yes & \yes & \res{99.1} & \res{67.4} & \res{94.2} & \res{86.9} & \res{80.2} \\

    \bottomrule
  \end{tabular}
  \caption{\textbf{Component contributions on the 60 Object, Goal, and Spatial tasks.}
  Held-out success (\%), with 20 tasks per suite and the same released starting library.
  Each row adds the marked components to the original protocol.
  Mean over the 60 tasks: Final as in the suite columns; First with each task's first attempt, before tasks below 80\% are rerun.}
  \label{tab:ablation}
\end{table}

\begin{figure*}[t]
  \centering
  \includegraphics[width=\textwidth]{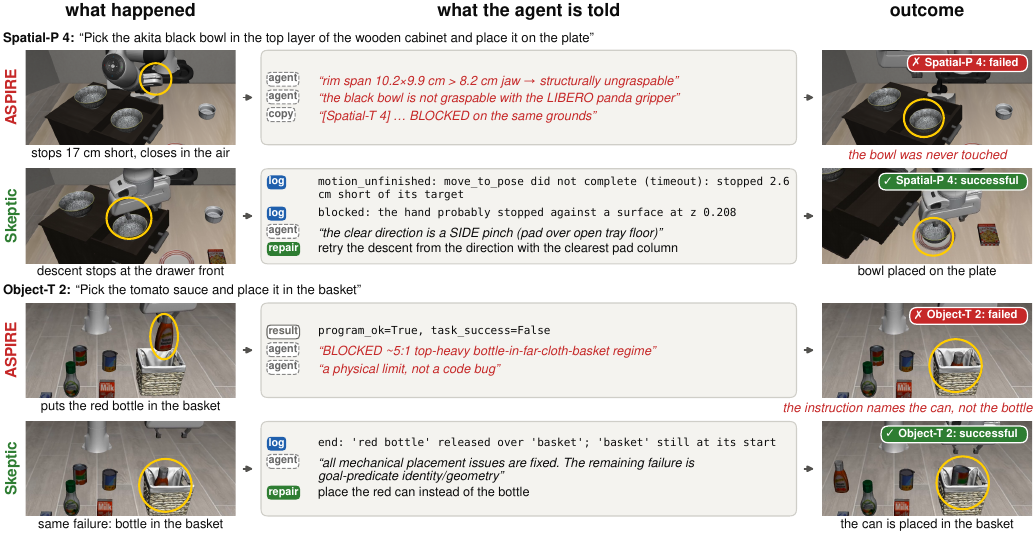}
  \caption{\textbf{Two more failures, with and without the event log.}
  \emph{Top (action events):} under \aspire, the hand stops short of the bowl and closes in the air; the agent concludes the bowl cannot be grasped.
  With \method, the log reports a descent stopped against a surface; the agent retries from a clear direction and completes the task.
  \emph{Bottom (end states):} the agent first places the red bottle in the basket and the task fails.
  Under \aspire, it attributes this to a physical limit of the placement.
  With \method, the end state shows the bottle released over the basket and the basket unmoved; the agent keeps the placement, checks which object the instruction names, and places the can.}
  \label{fig:examples}
\end{figure*}

\noindent\textbf{Component contributions.}
\Cref{tab:ablation} separates verified promotion from the two parts of the grounded event log.
Each alone improves on the original protocol (\res{70.1}\%): verified promotion reaches \res{81.3}\%, and the event log without verified promotion \res{80.8}\%.
Verified promotion with action events reaches \res{83.6}\%, and the full method, which also reports object end states, \res{86.9}\%.
Its scores are higher on all three suites, by \res{4.4}, \res{3.8}, and \res{1.6} points on Object, Goal, and Spatial, respectively.
Events describe what each action did; end states help locate the unfinished step when several actions must succeed in sequence.
\Cref{fig:examples} shows one task for each part of the log, both ending in a claim of impossibility under the original protocol.
In both, the log turns a failure that the original protocol reads as a physical limit into a specific repair, a new approach direction for the bowl and a different object for the basket, and held-out success rises from \res{0.0}\% to \res{45.0}\% and \res{98.0}\%.

\smallskip
\noindent\textbf{First attempts and reruns.}
Verified promotion and the event log act at different points.
Verified promotion alone raises success from \res{70.1}\% to \res{81.3}\%, but not first-attempt success (\res{54.0}\% against \res{61.3}\%):
\res{32} of the 60 tasks end their first attempt below 80\%, and \res{18} reach 80\% on the rerun, against \res{7} per campaign under \aspire.
Action events raise first-attempt success to \res{76.9}\%, and end states to \res{80.2}\%, leaving \res{13} tasks per campaign below 80\% after the first attempt.
On Goal, verified promotion alone reaches \res{68.2}\%, comparable to the full method (\res{67.4}\%).

\begin{figure}[t]
  \centering
  \includegraphics[width=\linewidth]{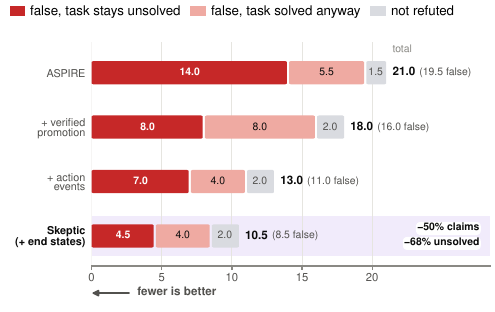}
  \caption{\textbf{Claims of impossibility and what they cost.}
  Tasks, out of the 60 Object, Goal, and Spatial tasks, with an attempt ending in a claim of impossibility; campaign means.
  A claim is false when any reported configuration reaches 80\% on the task; the task stays unsolved when it ends below 80\%.
  Percentages give the change from \aspire to \method.}
  \label{fig:audit}
\end{figure}

\smallskip
\noindent\textbf{Auditing failure judgments.}
\label{sec:audit}
We read the worker report and coordinator decision for every attempt ending below 80\% held-out success, and count a claim of impossibility when the terminal record asserts a required step cannot be done; explanations citing budget, chance, or a program bug do not count.
\Cref{fig:audit} gives the counts.

\smallskip
\noindent\textbf{Fewer tasks end on false claims.}
Under \aspire, impossibility claims occur on \res{21.0} tasks per campaign, of which \res{19.5} are false; \res{14.0} tasks finish below 80\% with a false claim.
With \method, these counts fall to \res{10.5}, \res{8.5}, and \res{4.5}.
The reduction appears in both campaigns: the lower-scoring \method campaign makes \res{11} false claims and leaves \res{7} unresolved, against \res{17} and \res{12} in the higher-scoring baseline.
Workers therefore make fewer false claims during discovery, beyond the restriction on what enters memory.
The plate-pushing and bottom-drawer tasks remain declared impossible in every campaign that attempts them.

\smallskip
\noindent\textbf{How the components affect claims.}
The log should improve diagnosis, while verified promotion controls which conclusions enter the shared library.
Verified promotion alone leaves false claims almost unchanged (\res{16} against \res{19.5} tasks) but nearly halves the tasks they cost (\res{8} against \res{14}): on eight tasks, the rerun reaches 80\% after the first attempt declared a step impossible.
The event log reduces the claims themselves: adding action events lowers them to \res{11.0}, and the full method to \res{8.5}; the two full-method campaigns have \res{6} and \res{11}.
Inheritance is rarer: three tasks in one baseline campaign are closed by the coordinator using a sibling task's conclusion, and no campaign records inheritance through the library.
Most remaining false claims concern Goal tasks, where the performance gap is also largest.

\subsection{Development cost}
\label{sec:t18}

\begin{figure}[t]
  \centering
  \includegraphics[width=\linewidth]{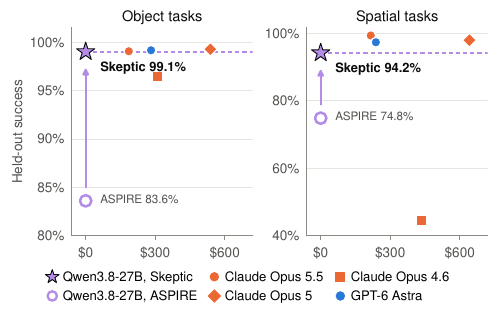}
  \caption{\textbf{Held-out success and API charges on the Object and Spatial tasks.}
  Each panel covers one suite (20 tasks).
  The horizontal axis is the API charge for these tasks at list prices, from the per-suite cost per task reported by \aspire~\cite{lu2026aspire}; the dashed line marks the success of \method, which incurs no API charge.}
  \label{fig:cost}
\end{figure}

\begin{figure}[t]
  \centering
  \includegraphics[width=\linewidth]{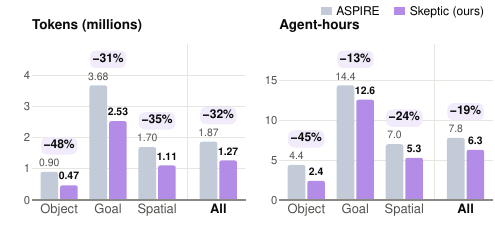}
  \caption{\textbf{Development resources per successful execution.}
  Generated tokens (millions) and agent-hours per 100 successful held-out episodes on the 60 Object, Goal, and Spatial tasks; All divides the totals by the total successes.
  Labels show changes relative to \aspire.
  Counts include workers, verifiers, and reruns; coordinator work and held-out evaluation are excluded.
  \method needs fewer resources per success on every suite: about a third fewer tokens and a fifth fewer agent-hours overall.}
  \label{fig:compute}
\end{figure}

\noindent\textbf{Resources per successful execution.}
\method uses fewer generated tokens and agent-hours per 100 successful held-out episodes on every suite (\cref{fig:compute}).
Across the 60 tasks, tokens decrease from 1.87M to 1.27M and agent-hours from 7.8 to 6.3.
On Goal tasks, tokens decrease from 3.68M to 2.53M.
The additional execution feedback is therefore accompanied by lower development resources relative to the number of successful held-out executions.

\smallskip
\noindent\textbf{Local inference.}
Both protocols run the same model locally and incur no API charges.
\Cref{fig:cost} compares them with the four proprietary models of \cref{tab:main} on the Object and Spatial suites.
On Object, \method raises success from \res{83.6}\% to \res{99.1}\%, matching the three strongest proprietary models (99.1--99.3\%), whose estimated API charges for the suite's 20 tasks range from \$186 to \$538.
On Spatial, it raises success from \res{74.8}\% to \res{94.2}\%, 3--5 points below these models and above Claude Opus 4.6 (44.5\%, \$434).

\section{Conclusion}
\label{sec:conclusion}

We studied agentic robot skill discovery with a compact open-weight model and identified a failure in how it learns from physical interaction.
An unsuccessful execution can become an unsupported judgment of physical impossibility, ending exploration and misleading subsequent tasks.
We introduced \method to improve how execution experience informs program repair and skill reuse.
Its grounded event log makes action outcomes and object end states explicit, while verified promotion requires source-program success and independent reproduction before skills enter the library.
With frozen weights and the same revision budget, \method improves held-out success by \res{14.9} points on LIBERO-Pro and \res{11.7} on Robosuite.
These results show that the feedback and memory mechanisms surrounding a compact model are consequential parts of robot skill discovery: they determine how failed interactions guide further attempts and which experience becomes reusable knowledge.

\smallskip
\noindent\textbf{Limitations.}
Our experiments use a single model, \qwen, on two simulated benchmarks.
In preliminary runs, five other open-weight models of similar size did not complete the original discovery loop, so \method is evaluated where that loop already runs; whether the gains carry over to other compact models and to real robots remains to be tested.
A gap to frontier proprietary models also remains (\cref{tab:main}).

\section*{LLM Usage Statement}
We used Claude (Anthropic) as an assistant in this work; no other LLM or generative AI tool was used.
It assisted with discussing ideas and related work, implementing and debugging code, preparing figures and tables, and writing and editing the text, including grammar.
In each of these, the authors made the research decisions and judgments: the problem, the method, the experimental design, and the conclusions are the authors' own.
This assistance is separate from the role of language models as the agents evaluated in our experiments (\cref{sec:setup}), which is the subject of this work.
We did not use AI tools to generate or modify the reported results: every number in the paper comes from logged experimental runs.
The authors reviewed and verified all AI-assisted content and take full responsibility for the paper.

\clearpage
\maketitlesupplementary
\appendix
\noindent This supplementary material is organized as follows.
\Cref{sec:supp-curation} presents the experiment underlying the distinction, drawn in the Introduction, between executing and discovering a strategy.
\Cref{sec:supp-false-negatives} lists the claims of impossibility identified by the audit of \cref{sec:audit}.
\Cref{sec:supp-per-task} reports the per-task results underlying \cref{tab:main}.
\Cref{sec:supp-recent} compares \method with two other agentic systems evaluated on the same tasks.
\Cref{sec:supp-failures} examines two tasks that a campaign with \method closes as blocked, and \cref{sec:supp-promotion} traces the skill library across two consecutive tasks under each protocol.
\Cref{sec:supp-library} presents entries of the learned skill libraries together with their admission records.
\Cref{sec:supp-open-models} reports the other open-weight models referred to in the Limitations.

\section{Execution versus curation}
\label{sec:supp-curation}
This experiment separates the ability of \qwen to execute a manipulation strategy from its ability to discover one.
In an initial campaign under the original protocol, conducted separately from the campaigns of \cref{tab:main}, \qwen attains 0 of 50 held-out successes on seven Goal tasks.
We compare three conditions for the same model on these seven tasks (\cref{tab:supp-curation}).

Removing every claim of impossibility from the model's own library, and requiring three distinct strategies before a task may be declared blocked, improves a single task and yields 9 of 350 held-out successes.
In contrast, a library curated by \opus on the same tasks yields 342 of 350.
The 27B model thus executes these tasks when a suitable strategy is provided, yet does not discover such a strategy under the original protocol.
Even together, library cleanup and the instruction to explore alternatives leave six tasks unsolved. \method provides evidence about the outcome of each execution.
In the campaigns of \cref{tab:main}, \method attains between \res{49.0}\% and \res{95.0}\% on five of the seven tasks, whereas the original protocol exceeds zero on one (\cref{tab:supp-per-task-goal}); Goal-P 5 and Goal-T 0 remain unsolved.

\begin{table}[t]
  \centering
  \footnotesize
  \setlength{\tabcolsep}{3pt}
  \resizebox{\ifdim\width>\columnwidth\columnwidth\else\width\fi}{!}{%
  \begin{tabular}{@{}l l c c c@{}}
    \toprule
    & & \multicolumn{3}{c@{}}{Skill library} \\
    \cmidrule(l){3-5}
    Task & Name & Own & Cleaned$^\dagger$ & Opus~5.5 \\
    \midrule
    Goal-P 0 & Open middle drawer & 0.0 & 0.0 & 100.0 \\
    Goal-P 5 & Push plate to stove & 0.0 & 0.0 & 90.0 \\
    Goal-P 7 & Turn on stove & 0.0 & 0.0 & 100.0 \\
    Goal-T 0 & Open bottom drawer & 0.0 & 0.0 & 100.0 \\
    Goal-T 1 & Plate on stove & 0.0 & 0.0 & 94.0 \\
    Goal-T 3 & Open top drawer+cream cheese & 0.0 & 0.0 & 100.0 \\
    Goal-T 4 & Plate on cabinet top & 0.0 & 18.0 & 100.0 \\
    \midrule
    Mean & & 0.0 & 2.6 & 97.7 \\
    \bottomrule
  \end{tabular}}
  \caption{\textbf{The 27B model executes strategies that it does not discover under the original protocol.}
  Held-out success (\%) of \qwen on seven Goal tasks (50 seeds per task, names as in \cref{tab:supp-per-task-goal}).
  $^\dagger$We remove or rewrite all claims of impossibility and require workers to try three distinct strategies before declaring a task blocked.
  The curated library includes code, so its score is an upper bound on the benefit of a better library.}
  \label{tab:supp-curation}
\end{table}

\section{False claims of impossibility}
\label{sec:supp-false-negatives}
\Cref{tab:supp-false-negatives} lists the claims of impossibility identified by the audit of \cref{sec:audit}, with which the coordinator closes a task under the original protocol.
The table comprises every conclusion in the report of the first original-protocol campaign that satisfies three conditions: it closes a task without a held-out measurement, it asserts that a required step cannot be performed, and it is refuted by a campaign of \cref{tab:main}.
We additionally include the claim illustrated in \cref{fig:examples}, which is recorded in the second campaign.
Two further claims of the first report, concerning Goal-P 5 and Goal-T 0, are not refuted, as no campaign solves these tasks.

Every listed task attains at least \res{90}\% held-out success in one campaign of \cref{tab:main}; every listed claim is therefore false under the criterion of \cref{fig:audit}.
The second campaign of the original protocol itself solves six of the tasks (Object-T 3, Object-T 8, Goal-P 0, Goal-P 8, Spatial-T 1, and Spatial-T 4), whereas on the remaining five only the campaigns with \method exceed 80\%.
Goal-P 3 and Goal-T 3 correspond to the drawer tasks of \cref{fig:teaser}, and Spatial-P 4, Spatial-T 4, and Object-T 2 to the tasks of \cref{fig:examples}.

The claims share a common structure.
Several cite a valid measurement, such as a displacement of 1.9\,cm or a rim span of 10.2\,cm, and err in the inference from that measurement to the infeasibility of the task.
Others rely on a measurement of the wrong object (Goal-T 1) or adopt the conclusion of a related task without failure evidence of their own (Goal-T 3 and Spatial-T 4).

\begin{table*}[t]
  \centering
  \showresultstrue
  \footnotesize
  \setlength{\tabcolsep}{5pt}
  \begin{tabular}{@{}l l l c@{}}
    \toprule
    Task & Name & Conclusion recorded by \aspire & Highest (\%) \\
    \midrule
    Object-T 2 & Tomato sauce & a top-heavy bottle in the far basket is a physical limit, not a code bug$^\ddagger$ & \res{100} \\
    Object-T 3 & Ketchup & the strict rest-pose predicate cannot be satisfied & \res{100} \\
    Object-T 8 & Salad dressing & the basket floor is beyond wrist reach & \res{100} \\
    Goal-P 0 & Open middle drawer & the hand cannot reach down to the drawer handle & \res{100} \\
    Goal-P 3 & Open top drawer+bowl & the top drawer stops after 1.9\,cm (mechanical stop) & \res{94} \\
    Goal-T 3 & Open top drawer+cream cheese & \emph{reuses the conclusion of Goal-P 3} & \res{94} \\
    Goal-P 8 & Bowl on plate & the program aborts on layouts it judges blocked & \res{100} \\
    Goal-T 1 & Plate on stove & the bowl (10.8\,cm) is wider than the jaw (8.2\,cm)$^\dagger$ & \res{96} \\
    Spatial-P 4 & In top drawer & the rim (10.2\,cm) is wider than the jaw: ungraspable & \res{90} \\
    Spatial-T 4 & On top of wooden cabinet & \emph{reuses the conclusion of Spatial-P 4} & \res{96} \\
    Spatial-T 1 & Next to cookie box & the placement predicate is stricter than any achievable centering & \res{100} \\
    \bottomrule
  \end{tabular}
  \caption{\textbf{False claims of impossibility recorded under the original protocol.}
  Each row gives the conclusion with which the 27B coordinator closes a task, paraphrased from its campaign reports, and the highest held-out success attained on that task by a single campaign of \cref{tab:main}.
  Task names follow \cref{tab:supp-per-task-object,tab:supp-per-task-goal,tab:supp-per-task-spatial}.
  $^\dagger$The instruction refers to the plate, whereas the coordinator measured the bowl named in the task identifier.
  $^\ddagger$Recorded in the second campaign; the first campaign deferred the task.}
  \label{tab:supp-false-negatives}
\end{table*}

\section{Per-task results}
\label{sec:supp-per-task}

\Cref{tab:supp-per-task-object,tab:supp-per-task-goal,tab:supp-per-task-spatial,tab:supp-per-task-long} report the per-task results underlying \cref{tab:main}.
For both \aspire and \method on LIBERO-Pro, we set the reasoning effort of \qwen to \texttt{medium}.
Each entry is the mean over two campaigns.

\begin{figure*}[t]
  \centering
  \includegraphics[width=\textwidth]{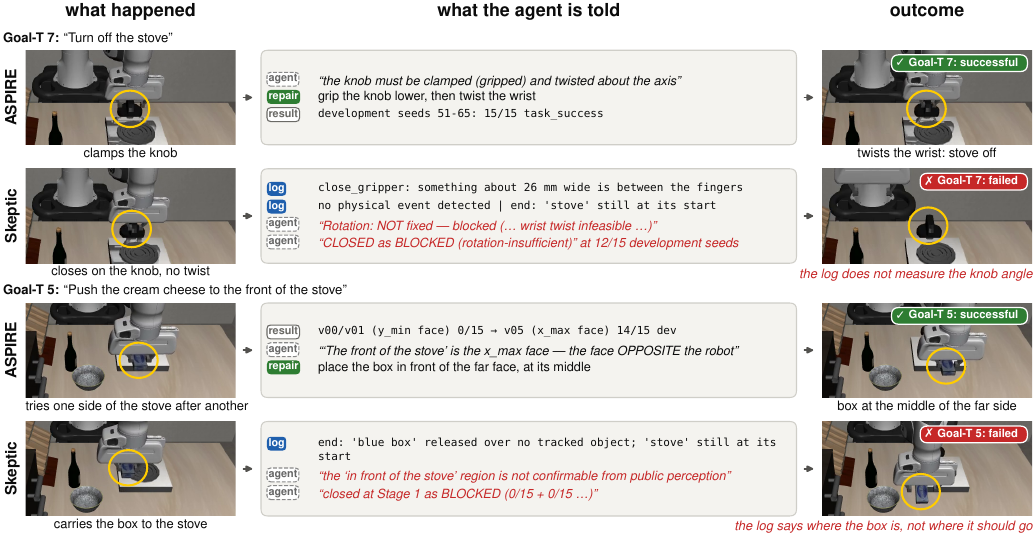}
  \caption{\textbf{Two tasks solved under the original protocol and closed as blocked by a campaign with \method.}
  \emph{Top:} under \aspire, the agent clamps the stove knob and rotates the wrist.
  With \method, the program closes the gripper on the knob without a rotation and succeeds on 12 of 15 development seeds; the log reports the grasp and an unmoved stove but no measurement of the knob angle, and the agent declares the rotation blocked.
  \emph{Bottom:} under \aspire, the agent evaluates the sides of the stove successively and identifies the one designated by the instruction.
  With \method, the log reports the release position of the box, and the agent concludes that the goal region cannot be identified.}
  \label{fig:failures}
\end{figure*}

On the 60 Object, Goal, and Spatial tasks, \method attains higher success on \res{36}, equal success on \res{13}, and lower success on \res{11}.
The largest decreases occur on Goal-T 7 (99.0 to \res{49.0}) and Goal-T 5 (43.0 to \res{0.0}).
On Goal-T 7, the second campaign with \method closes the task as blocked without a held-out measurement, although its program succeeds on 12 of 15 development seeds.
On Goal-T 5, both campaigns with \method attain 0, whereas one campaign of the original protocol attains \res{86}\%.
\Cref{fig:failures} examines both tasks.

On the 20 Long tasks (\cref{tab:supp-per-task-long}), \method attains higher success on \res{8}, equal success on \res{5}, and lower success on \res{7}; the improvement is concentrated in a few tasks, most notably Long-T 7, Long-P 4, and Long-T 5.
Four tasks that require closing a drawer or a microwave after placing an object (Long-P 3, Long-P 9, Long-T 3, and Long-T 9) attain 0 in every campaign of \cref{tab:main}.

\begin{table}[t]
  \centering
  \showresultstrue
  \footnotesize
  \setlength{\tabcolsep}{3pt}
  \begin{tabular}{@{}l l c >{\columncolor{blue!4}}c@{}}
    \toprule
    Task & Object & \aspire & \method \\
    \midrule
    \multicolumn{4}{@{}l}{\textit{Position perturbation}} \\
    Object-P 0 & Alphabet soup & \res{100.0} & \res{100.0} \\
    Object-P 1 & Cream cheese & \res{100.0} & \res{100.0} \\
    Object-P 2 & Salad dressing & \res{44.0} & \res{100.0} \\
    Object-P 3 & BBQ sauce & \res{97.0} & \res{100.0} \\
    Object-P 4 & Ketchup & \res{50.0} & \res{96.0} \\
    Object-P 5 & Tomato sauce & \res{98.0} & \res{100.0} \\
    Object-P 6 & Butter & \res{100.0} & \res{100.0} \\
    Object-P 7 & Milk & \res{99.0} & \res{100.0} \\
    Object-P 8 & Chocolate pudding & \res{100.0} & \res{100.0} \\
    Object-P 9 & Orange juice & \res{92.0} & \res{100.0} \\
    \cmidrule(l){3-4}
    Mean & & \res{88.0} & \res{99.6} \\
    \midrule
    \multicolumn{4}{@{}l}{\textit{Task perturbation}} \\
    Object-T 0 & Cream cheese & \res{100.0} & \res{100.0} \\
    Object-T 1 & Alphabet soup & \res{98.0} & \res{100.0} \\
    Object-T 2 & Tomato sauce & \res{0.0} & \res{98.0} \\
    Object-T 3 & Ketchup & \res{50.0} & \res{100.0} \\
    Object-T 4 & Milk & \res{100.0} & \res{91.0} \\
    Object-T 5 & BBQ sauce & \res{100.0} & \res{97.0} \\
    Object-T 6 & Orange juice & \res{94.0} & \res{99.0} \\
    Object-T 7 & Butter & \res{100.0} & \res{100.0} \\
    Object-T 8 & Salad dressing & \res{50.0} & \res{100.0} \\
    Object-T 9 & Chocolate pudding & \res{100.0} & \res{100.0} \\
    \cmidrule(l){3-4}
    Mean & & \res{79.2} & \res{98.5} \\
    \bottomrule
  \end{tabular}
  \caption{\textbf{Per-task held-out success (\%) on LIBERO-Pro Object.}
  \qwen under the original protocol (\aspire) and with \method; each Mean equals the corresponding entry of \cref{tab:main}.
  Every task places the listed object in the basket.}
  \label{tab:supp-per-task-object}
\end{table}

\begin{table}[t]
  \centering
  \showresultstrue
  \footnotesize
  \setlength{\tabcolsep}{3pt}
  \begin{tabular}{@{}l l c >{\columncolor{blue!4}}c@{}}
    \toprule
    Task & Instruction (abbreviated) & \aspire & \method \\
    \midrule
    \multicolumn{4}{@{}l}{\textit{Position perturbation}} \\
    Goal-P 0 & Open middle drawer & \res{45.0} & \res{50.0} \\
    Goal-P 1 & Bowl on stove & \res{56.0} & \res{88.0} \\
    Goal-P 2 & Wine on cabinet top & \res{96.0} & \res{100.0} \\
    Goal-P 3 & Open top drawer+bowl & \res{0.0} & \res{65.0} \\
    Goal-P 4 & Bowl on cabinet top & \res{98.0} & \res{96.0} \\
    Goal-P 5 & Push plate to stove & \res{0.0} & \res{0.0} \\
    Goal-P 6 & Cream cheese in bowl & \res{81.0} & \res{54.0} \\
    Goal-P 7 & Turn on stove & \res{0.0} & \res{60.0} \\
    Goal-P 8 & Bowl on plate & \res{49.0} & \res{96.0} \\
    Goal-P 9 & Wine on rack & \res{84.0} & \res{100.0} \\
    \cmidrule(l){3-4}
    Mean & & \res{50.9} & \res{70.9} \\
    \midrule
    \multicolumn{4}{@{}l}{\textit{Task perturbation}} \\
    Goal-T 0 & Open bottom drawer & \res{0.0} & \res{0.0} \\
    Goal-T 1 & Plate on stove & \res{0.0} & \res{95.0} \\
    Goal-T 2 & Wine in bowl & \res{94.0} & \res{81.0} \\
    Goal-T 3 & Open top drawer+cream cheese & \res{0.0} & \res{68.0} \\
    Goal-T 4 & Plate on cabinet top & \res{0.0} & \res{49.0} \\
    Goal-T 5 & Push cream cheese to stove & \res{43.0} & \res{0.0} \\
    Goal-T 6 & Wine in bowl & \res{98.0} & \res{98.0} \\
    Goal-T 7 & Turn off stove & \res{99.0} & \res{49.0} \\
    Goal-T 8 & Wine on plate & \res{98.0} & \res{99.0} \\
    Goal-T 9 & Cream cheese on rack & \res{96.0} & \res{99.0} \\
    \cmidrule(l){3-4}
    Mean & & \res{52.8} & \res{63.8} \\
    \bottomrule
  \end{tabular}
  \caption{\textbf{Per-task held-out success (\%) on LIBERO-Pro Goal.} As in \cref{tab:supp-per-task-object}.}
  \label{tab:supp-per-task-goal}
\end{table}

\begin{table}[t]
  \centering
  \showresultstrue
  \footnotesize
  \setlength{\tabcolsep}{3pt}
  \begin{tabular}{@{}l l c >{\columncolor{blue!4}}c@{}}
    \toprule
    Task & Bowl location & \aspire & \method \\
    \midrule
    \multicolumn{4}{@{}l}{\textit{Position perturbation}} \\
    Spatial-P 0 & Betw. plate+ramekin & \res{94.0} & \res{94.0} \\
    Spatial-P 1 & Next to ramekin & \res{42.0} & \res{96.0} \\
    Spatial-P 2 & From table center & \res{100.0} & \res{98.0} \\
    Spatial-P 3 & On cookie box & \res{0.0} & \res{96.0} \\
    Spatial-P 4 & In top drawer & \res{0.0} & \res{45.0} \\
    Spatial-P 5 & On ramekin & \res{50.0} & \res{94.0} \\
    Spatial-P 6 & Next to cookie box & \res{98.0} & \res{98.0} \\
    Spatial-P 7 & On stove & \res{99.0} & \res{94.0} \\
    Spatial-P 8 & Next to plate & \res{96.0} & \res{100.0} \\
    Spatial-P 9 & On wooden cabinet & \res{93.0} & \res{100.0} \\
    \cmidrule(l){3-4}
    Mean & & \res{67.2} & \res{91.5} \\
    \midrule
    \multicolumn{4}{@{}l}{\textit{Task perturbation}} \\
    Spatial-T 0 & Not betw. plate+ramekin & \res{47.0} & \res{95.0} \\
    Spatial-T 1 & Next to cookie box & \res{50.0} & \res{100.0} \\
    Spatial-T 2 & Next to plate & \res{89.0} & \res{92.0} \\
    Spatial-T 3 & On top of cabinet & \res{98.0} & \res{100.0} \\
    Spatial-T 4 & On top of wooden cabinet & \res{48.0} & \res{89.0} \\
    Spatial-T 5 & On cookie box & \res{100.0} & \res{100.0} \\
    Spatial-T 6 & On stove & \res{99.0} & \res{97.0} \\
    Spatial-T 7 & On top of cabinet & \res{99.0} & \res{100.0} \\
    Spatial-T 8 & Next to ramekin & \res{97.0} & \res{99.0} \\
    Spatial-T 9 & On stove & \res{97.0} & \res{96.0} \\
    \cmidrule(l){3-4}
    Mean & & \res{82.4} & \res{96.8} \\
    \bottomrule
  \end{tabular}
  \caption{\textbf{Per-task held-out success (\%) on LIBERO-Pro Spatial.} As in \cref{tab:supp-per-task-object}. Every task picks the black bowl from the listed location and places it on the plate.}
  \label{tab:supp-per-task-spatial}
\end{table}

\begin{table}[t]
  \centering
  \showresultstrue
  \footnotesize
  \setlength{\tabcolsep}{3pt}
  \begin{tabular}{@{}l l c >{\columncolor{blue!4}}c@{}}
    \toprule
    Task & Instruction (abbreviated) & \aspire & \method \\
    \midrule
    \multicolumn{4}{@{}l}{\textit{Position perturbation}} \\
    Long-P 0 & Soup + sauce $\rightarrow$ basket & \res{23.0} & \res{11.0} \\
    Long-P 1 & Cheese + butter $\rightarrow$ basket & \res{85.0} & \res{86.0} \\
    Long-P 2 & Stove on + moka pot & \res{100.0} & \res{100.0} \\
    Long-P 3 & Bowl $\rightarrow$ drawer, close & \res{0.0} & \res{0.0} \\
    Long-P 4 & Mugs $\rightarrow$ L/R plates & \res{5.0} & \res{63.0} \\
    Long-P 5 & Book $\rightarrow$ caddy back & \res{50.0} & \res{31.0} \\
    Long-P 6 & Mug $\rightarrow$ plate, pudding right & \res{0.0} & \res{43.0} \\
    Long-P 7 & Soup + cheese $\rightarrow$ basket & \res{92.0} & \res{64.0} \\
    Long-P 8 & Both moka pots $\rightarrow$ stove & \res{12.0} & \res{40.0} \\
    Long-P 9 & Mug $\rightarrow$ microwave, close & \res{0.0} & \res{0.0} \\
    \cmidrule(l){3-4}
    Mean & & \res{36.7} & \res{43.8} \\
    \midrule
    \multicolumn{4}{@{}l}{\textit{Task perturbation}} \\
    Long-T 0 & Cheese + sauce $\rightarrow$ basket & \res{50.0} & \res{43.0} \\
    Long-T 1 & Soup + butter $\rightarrow$ basket & \res{90.0} & \res{87.0} \\
    Long-T 2 & Stove on + pan & \res{85.0} & \res{100.0} \\
    Long-T 3 & Bottle $\rightarrow$ drawer, close & \res{0.0} & \res{0.0} \\
    Long-T 4 & Mugs $\rightarrow$ L/R plates, swapped & \res{78.0} & \res{51.0} \\
    Long-T 5 & Cup $\rightarrow$ caddy back & \res{42.0} & \res{96.0} \\
    Long-T 6 & Red mug $\rightarrow$ plate, pudding & \res{38.0} & \res{44.0} \\
    Long-T 7 & Ketchup + cheese $\rightarrow$ basket & \res{17.0} & \res{97.0} \\
    Long-T 8 & Left moka pot $\rightarrow$ stove & \res{100.0} & \res{96.0} \\
    Long-T 9 & Mug $\rightarrow$ microwave, close & \res{0.0} & \res{0.0} \\
    \cmidrule(l){3-4}
    Mean & & \res{50.0} & \res{61.4} \\
    \bottomrule
  \end{tabular}
  \caption{\textbf{Per-task held-out success (\%) on LIBERO-Pro Long.} As in \cref{tab:supp-per-task-object}.}
  \label{tab:supp-per-task-long}
\end{table}

\FloatBarrier

\section{Comparison with other agentic systems}
\label{sec:supp-recent}
\Cref{tab:supp-recent} compares \method with World Action Agent (WAA)~\cite{zhang2026waa} and SPARK~\cite{grant2026spark} on the same 60 LIBERO-Pro tasks. All three systems use frozen model weights.
WAA reports six evaluation episodes per task. SPARK and \method each report 50 per task.

\noindent\begin{minipage}{\columnwidth}
  \centering
  \showresultstrue
  \footnotesize
  \setlength{\tabcolsep}{5pt}
  \begin{tabular}{@{}l cc cc cc c@{}}
    \toprule
    & \multicolumn{2}{c}{Object} & \multicolumn{2}{c}{Goal} & \multicolumn{2}{c}{Spatial} & \\
    \cmidrule(lr){2-3}\cmidrule(lr){4-5}\cmidrule(lr){6-7}
    Method & Pos & Task & Pos & Task & Pos & Task & Mean \\
    \midrule
    WAA~\cite{zhang2026waa} & 95.0 & 93.3 & 63.3 & 48.3 & 80.0 & 73.3 & 75.6 \\
    SPARK~\cite{grant2026spark} & 86.4 & 82.4 & \textbf{71.6} & 61.8 & 65.4 & 83.0 & 75.1 \\
    \method & \res{\textbf{99.6}} & \res{\textbf{98.5}} & \res{70.9} & \res{\textbf{63.8}} & \res{\textbf{91.5}} & \res{\textbf{96.8}} & \res{\textbf{86.9}} \\
    \bottomrule
  \end{tabular}
  \captionof{table}{\textbf{Comparison with other agentic systems on LIBERO-Pro.}
  Success (\%) on the 60 Object, Goal, and Spatial tasks.
  Results of WAA and SPARK are taken from the respective papers; bold marks the best result in each column.}
  \label{tab:supp-recent}
\end{minipage}\par

\smallskip
\noindent\textbf{Results.}
Across the six shared LIBERO-Pro settings, \method achieves \res{86.9}\% mean success, compared with 75.6\% for WAA and 75.1\% for SPARK under their respective protocols.
\method leads in five of the six settings. Its largest margin over SPARK is on Spatial-Pos (\res{26.1} percentage points).

\smallskip
\noindent\textbf{Skill acquisition and use.}
WAA learns its skill library from LIBERO-90 only (text seed skills enriched with one expert trajectory per source task) and freezes it before evaluation. During an episode, Gemini 3.7 Flash~\cite{google2026gemini37flash} uses perception and motion tools to rehearse proposed actions and adjust them before execution.
SPARK uses a fixed library of author-reviewed skills. Given the task instruction and current observation, its 4-bit \qwen planner composes these skills into a behavior tree and can replan if local recovery fails.
\method revises each task's program using execution feedback on development seeds, retains verified skills for later tasks, and runs the selected program on held-out seeds (\cref{sec:setup}).

\smallskip
\noindent\textbf{Privileged information.}
SPARK reads the simulator's task-success signal after each attempt to decide whether to retry. When it reads that signal only at the end of the episode, its mean success is 67.5\%.
WAA uses a simulator-provided scene point cloud. Reconstructing the point cloud from RGB-D views yields 71.1\% mean success.
During held-out evaluation, \method uses depth images for perception and reads task success only at the end of the episode.

\section{Failure cases}
\label{sec:supp-failures}
\Cref{fig:failures} examines the two largest decreases in \cref{tab:supp-per-task-goal}: tasks that the original protocol solves and that a campaign with \method closes as blocked.

Goal-T 7 requires turning a stove knob.
Under the original protocol, the program clamps the knob and rotates the wrist, succeeding on all 50 held-out seeds.
With \method, the program closes the gripper on the knob without a rotation and succeeds on 12 of 15 development seeds.
For the three failed seeds, the log reports that the gripper holds an object approximately 26\,mm wide and that the stove has not moved; it provides no measurement of the knob angle.
The worker concludes that the rotation cannot be completed, and the coordinator closes the task without a held-out evaluation.

In Goal-T 5, success depends on which side of the stove the instruction designates as its front.
Under the original protocol, the agent evaluates the sides successively and finds that the side facing away from the robot satisfies the task (43 of 50 held-out seeds); its program carries the box to this location rather than pushing it.
With \method, the log reports the release position of the box, 3.2\,cm outside the footprint of the stove, but cannot indicate which region the instruction requires.
After placements at several sides fail, the worker concludes that the goal region cannot be identified.

Both cases delineate the scope of the grounded event log.
The log measures hand motion, jaw opening, and depth changes of segmented objects; it therefore captures neither the state of an articulated part that remains in place nor the goal region implied by an instruction.
In both cases, the agent again converts missing evidence into a claim of impossibility, a behavior that \method reduces but does not eliminate (\cref{fig:audit}).

\section{Skill admission across two tasks}
\label{sec:supp-promotion}
\begin{figure*}[t]
  \centering
  \includegraphics[width=\textwidth]{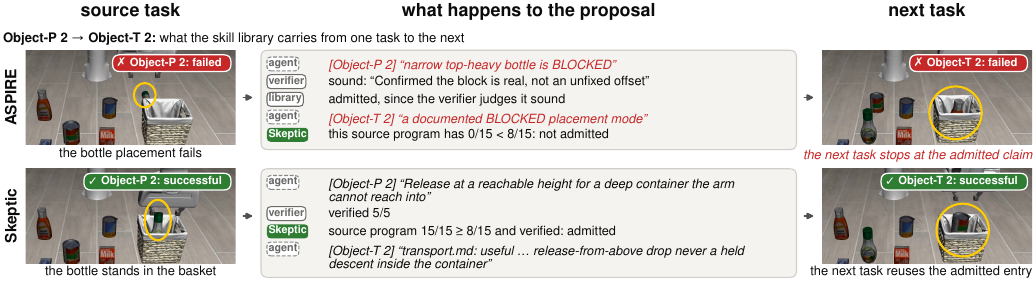}
  \caption{\textbf{Content carried by the skill library from one task to the next.}
  \emph{Top:} under \aspire, a program that fails on all 15 development seeds yields the claim that the bottle cannot be placed; the verifier reproduces the failure and judges the claim sound, the claim is admitted, and the subsequent task cites it and fails.
  \emph{Bottom:} with \method, the program for the same task succeeds on all 15 development seeds; its placement strategy is verified and admitted, and the subsequent task reuses it and succeeds.}
  \label{fig:promotion}
\end{figure*}
\begin{figure*}[t]
  \centering
  \includegraphics[width=\textwidth]{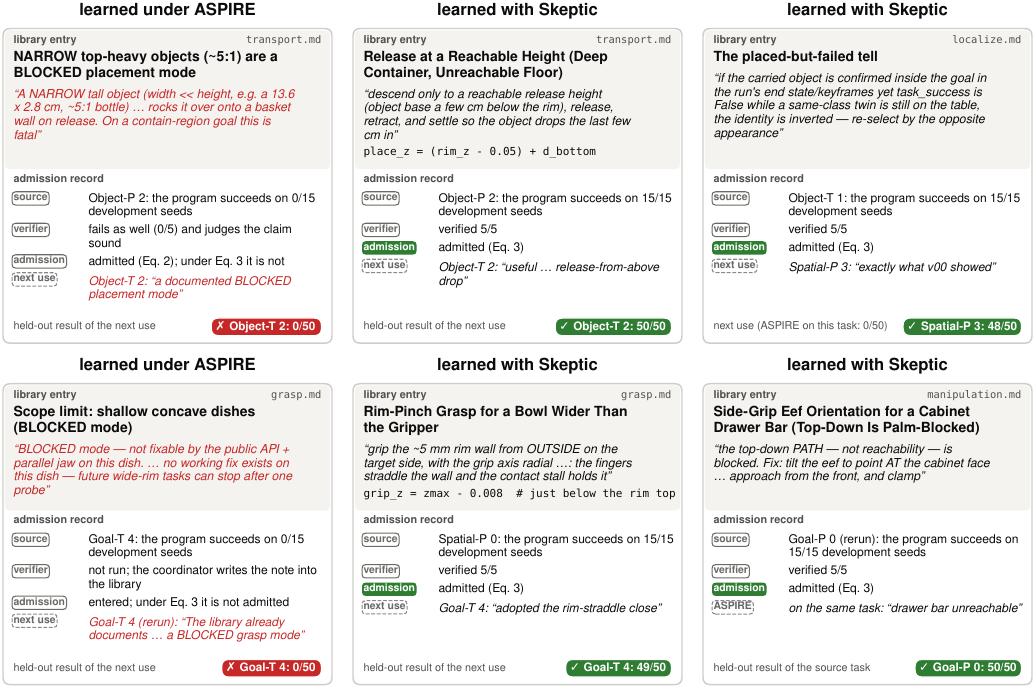}
  \caption{\textbf{Six entries of the learned skill libraries with their admission records.}
  Entry text is quoted from the final library of a campaign.
  In each row, the left entry is learned under \aspire: a claim of impossibility originating from a program without a development success, which a later attempt reads before failing.
  The middle entry is learned with \method and used by the same task, which subsequently succeeds.
  The right entries are two further entries admitted by \method: a rule triggered by the end state reported by the grounded log (top), and a grasp for a task that \aspire closed as unreachable (bottom).}
  \label{fig:library}
\end{figure*}

\Cref{fig:promotion} traces the skill library from Object-P 2 to Object-T 2, the subsequent task in the same scene, under each protocol.
Under the original protocol, the program for Object-P 2 succeeds on 0 of 15 development seeds, and the worker proposes that a narrow, top-heavy bottle cannot be placed in the basket.
The verifier follows the proposal, likewise fails, and judges the proposal sound.
Reproducing a failure, however, cannot distinguish a step that is infeasible from one that neither agent has accomplished.
The claim is admitted under \cref{eq:aspire-admit}; the worker of Object-T 2 cites it, and the task attains 0 of 50 held-out successes.
The claim is false: Object-P 2 attains \res{88}\% in the other campaign of the original protocol and \res{100}\% in both campaigns with \method.
Under \cref{eq:admit}, the claim would not be admitted, since its source program falls below the threshold.

With \method, the program for Object-P 2 succeeds on all 15 development seeds, and its placement strategy is verified and admitted.
The worker of Object-T 2 reuses this strategy, and the task succeeds on all 50 held-out seeds.

The two rows differ in more than the admission rule.
With \method, Object-P 2 is solved with the aid of the event log, and the rule subsequently ensures that the admitted entry originates from a program that completes its task; \cref{tab:ablation} isolates the two components.
Across the two campaigns with \method, the campaign reports mark six attempts as ineligible for promotion because their source programs fall below the threshold.

\section{Skill-library entries}
\label{sec:supp-library}
\Cref{fig:library} presents six entries of the final skill libraries, each accompanied by the record that \cref{eq:admit} associates with it: the development success of its source program, the verifier's result, and the outcome following its admission.

Each row begins with an entry learned under the original protocol, followed by an entry learned with \method and used by the same task.
In the top row, both entries originate from Object-P 2 (\cref{fig:promotion}): a program that never succeeds yields the claim that the placement is infeasible, whereas a program that succeeds on every development seed yields the release strategy that Object-T 2 reuses.
In the bottom row, the task is Goal-T 4, which requires grasping a wide, shallow plate.
Under the original protocol, the program succeeds on no development seed, and the coordinator records in the library, without a verifier run, that the grasp is infeasible and that subsequent tasks involving wide rims may stop after a single probe; the rerun of the task reads this note and again attains 0 of 50 held-out successes.
With \method, the library contains a rim-pinch grasp admitted from Spatial-P 0; the worker of Goal-T 4 adopts it, and the task attains \res{98}\% in that campaign.
Subsequent workers mark this entry as useful in \res{23} attempts and as misleading in \res{2}.

The remaining two entries illustrate further content admitted by \method.
The first is a diagnostic rule triggered by an observation of the grounded log: an object that the end state locates inside the goal while the task still fails indicates that the wrong object was selected.
It is admitted from Object-T 1 and applied in Spatial-P 3, which attains \res{96}\% in the same campaign, compared with 0 in both campaigns of the original protocol.
The second is a side grip for a drawer handle, admitted from the rerun of Goal-P 0, a task that one campaign of the original protocol closed on the grounds that the handle could not be reached.

\begin{table}[t]
  \centering
  \showresultstrue
  \footnotesize
  \setlength{\tabcolsep}{3pt}
  \resizebox{\ifdim\width>\columnwidth\columnwidth\else\width\fi}{!}{%
  \begin{tabular}{@{}l c c c c@{}}
    \toprule
    Model & Tasks & Episodes & Errors & Held-out \\
    \midrule
    Gemma 4 26B-A4B & \res{2\,/\,0} & \res{474} & \res{98.7} & \res{0/250} \\
    Gemma 4 31B & \res{5\,/\,4} & \res{1151} & \res{59.9} & \res{0/250} \\
    Devstral Small 2 24B & \res{5\,/\,5} & \res{1420} & \res{58.2} & \res{0/250} \\
    Muse Glimmer 30B & \res{5\,/\,2} & \res{1925} & \res{40.7} & \res{0/250} \\
    Nemotron 3.5 Super VL 120B-A12B & \res{5\,/\,4} & \res{687} & \res{63.6} & \res{0/250} \\
    \bottomrule
  \end{tabular}}
  \caption{\textbf{Other open-weight models as the agent under the original protocol.}
  Results on five LIBERO-Pro tasks, first attempts only.
  Tasks gives the number of tasks started and, after the slash, the number that completed a held-out evaluation.
  Episodes is the number of development episodes, none of which is successful, and Errors the percentage that terminate with a program error.
  Held-out counts successes out of 250 scheduled episodes per model (five tasks, 50 seeds each). Unrun episodes count as zero successes.}
  \label{tab:supp-open-models}
\end{table}

\section{Other open-weight models}
\label{sec:supp-open-models}
The Limitations paragraph notes that five other open-weight models did not complete the original discovery loop.
\Cref{tab:supp-open-models} reports these experiments.
Each model acts as coordinator, worker, and verifier under the original protocol on five LIBERO-Pro tasks (Spatial-P 0 and 2, Spatial-T 0 and 2, and Goal-P 4).
We consider first attempts only and initialize every run with the skill library used for the campaigns of \cref{tab:main}.
A run terminates once every task has either a held-out evaluation or has exhausted the budget of eight program versions without a successful episode, or after three hours without progress.

None of the five models completes a successful episode, neither among the \res{5657} development episodes nor in any held-out evaluation.
Program errors terminate \res{41}\% to \res{99}\% of the development episodes.
The models further depart from the protocol in distinct ways.
Gemma 4 26B-A4B~\cite{gemma2026gemma4} dispatches workers for two tasks and then stops.
Devstral Small 2~\cite{mistral2025devstral} interprets program completion as task success and declares the campaign complete.
Muse Glimmer~\cite{muse2026glimmer} and Nemotron 3.5 Super VL~\cite{nvidia2026nemotron35} continue to produce program versions well beyond the budget, up to \res{61} and \res{39} versions for a single task, without reporting a failure mode.
Gemma 4 31B~\cite{gemma2026gemma4} terminates after three consecutive sessions shorter than ten minutes.

Two models of the Qwen family provide a reference.
Under identical conditions, Qwen3.6-27B~\cite{qwen2026qwen36} solves one of the five tasks, using program versions beyond the budget, and reaches \res{36} of 250 held-out successes (\res{14.4}\%).
\qwen attains \res{85.6}\% on the same tasks under the original protocol in the campaigns of \cref{tab:main} (\cref{tab:supp-per-task-goal,tab:supp-per-task-spatial}).

These observations delimit the scope of our study.
\method modifies what a functioning discovery loop learns from its failures; we therefore evaluate it on the compact model for which this loop functions.

\end{document}